\documentclass{article}

\usepackage[preprint]{neurips_2026}

\usepackage[utf8]{inputenc} 
\usepackage[T1]{fontenc}    
\usepackage{hyperref}       
\usepackage{url}            
\usepackage{booktabs}       
\usepackage{amsfonts}       
\usepackage{nicefrac}       
\usepackage{microtype}      
\usepackage{xcolor}         

\usepackage{algorithm}
\usepackage{algorithmic}   

\newcommand{\model}{\textsc{ReFree-S2V}}

\definecolor{RDcolor}{rgb}{0.5, 0.1, 0.8}

\definecolor{bronze}{rgb}{1,1,0.6}
\definecolor{silver}{rgb}{0.969,0.796,0.600}
\definecolor{gold}{rgb}{0.941,0.592,0.600}

\definecolor{RDcolor}{rgb}{0.5, 0.1, 0.8}

\usepackage{graphicx}
\usepackage{amsmath}

\usepackage{booktabs} 
\usepackage{siunitx}  
\usepackage{etoolbox} 

\usepackage[table]{xcolor} 

\definecolor{TopResultColor}{RGB}{120, 150, 200}

\colorlet{first}{TopResultColor!30}
\colorlet{second}{TopResultColor!20}
\colorlet{third}{TopResultColor!10}
\robustify\bfseries

\title{\model: Towards Realistic Co-Speech Video Generation via Reward-Free RL and Multilevel Speech Guidance} 

\author{%
  \textbf{Salaheldin Mohamed} \\
  T\'el\'ecom Paris, Institut Polytechnique de Paris \\
  \and
  \textbf{M. Hamza Mughal} \\
  Max Planck Institute for Informatics \\
  \and
  \textbf{Rishabh Dabral} \\
  Max Planck Institute for Informatics \\
  \and
  \textbf{Christian Theobalt} \\
  Max Planck Institute for Informatics \\
}

\begin{document}

\maketitle

\begin{figure}[H]
  \centering
  \includegraphics[width=0.9\textwidth]{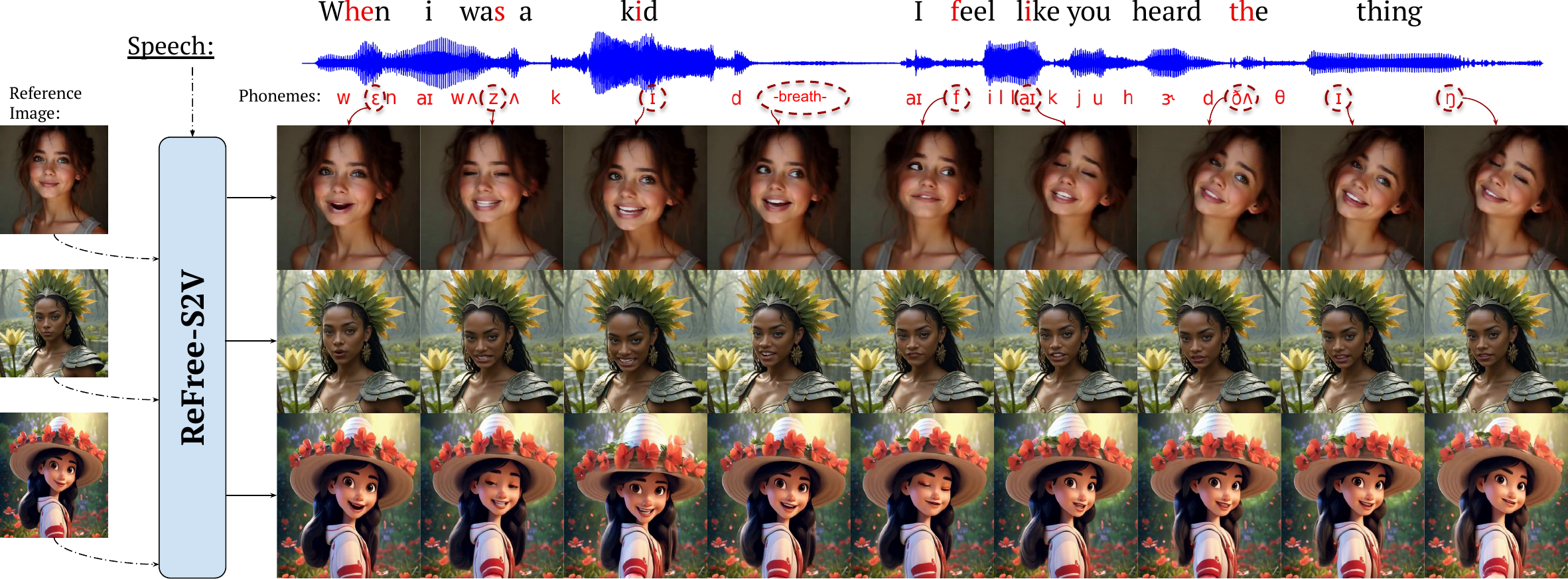}
  \caption{We present a speech-to-video generation framework that produces speech-synchronized, natural-looking portrait animations. Our approach produces accurate lip movements corresponding to speech's phonetic structure, and diverse and natural facial expressions and head movements.}  \label{fig:teaser}
\end{figure}

\footnotetext{Credit: initial first frame images are from \texttt{freepik.com}.}

\begin{abstract}
Speech-driven talking character animation seeks to generate life-like portrait videos that convey natural conversation behavior, aligning facial motion with spoken audio. Although recent advances in video generation have substantially improved realism in video-based animation, achieving both accurate lip articulation and expressive behavior remains challenging. Existing approaches typically trade off precise phoneme-to-lip synchronization against dynamic facial expressions and head motion, yielding animations that are either accurate yet rigid, or expressive but poorly synchronized. We address this challenge by proposing ReFree-S2V, a flow-matching speech-to-portrait animation framework that builds upon a pretrained video generation model to achieve fine-grained speech articulation and high-level expressive cues in speech-driven portrait animation. This model introduces a multi-level speech representation capturing phonetic and prosodic information at both local and global granularities. These representations are selectively injected into transformer blocks via learnable level selectors, enabling both accurate lip synchronization and natural expressive motion. To achieve natural head movements, we further introduce a novel reward-free reinforcement learning scheme into flow-matching training to discourage perceptually implausible motion without relying on handcrafted synchronization metrics or reward models, or the high cost of human preference annotation. Extensive experiments demonstrate that ReFree-S2V achieves state-of-the-art performance, significantly outperforming existing methods in both quantitative lip-sync accuracy and qualitative human evaluations of naturalness and expressivity.

\end{abstract}



\section{Introduction}
%
%
%
Virtual character animation has been a long-studied problem in computer vision and computer graphics. 
%
More recently, large video generation models~\cite{wan, openai2024video, yang2024cogvideox} have spurred renewed interest in animating life-like human characters, directly, in video. 
Some of these models~\cite{gao2025wan} aim not only to generate human videos but also to align the animation with underlying speech to achieve talking character animation. 
Such speech-driven video generation enables a variety of applications, ranging from content creation, video conversational agents, and human-computer interaction.  %
%
%
Given a reference portrait image and an input speech, the primary goal of talking character animation in video is to animate a life-like virtual character.
To achieve this, every framework must address two main challenges: 
(1) speech-to-lip synchronization, which entails that lip movements not only match the phonetic structure of speech but also look perceptually natural,
and 
(2) human-like expressivity, which requires generating dynamic and natural facial expressions with head motions driven by prosodic stress and semantic context in speech.
%
%
\par
%
%
%
%
Recent diffusion-based works~\cite{hallo3, sonic, fantasy} tackle speech-to-portrait animation using data-driven generative approaches. These methods typically build upon pretrained video transformers~\cite{yang2024cogvideox, blattmann2023stable} and rely on supervised fine-tuning (SFT) over large-scale datasets~\cite{fantasy, hallo3} to learn expressive speech-to-portrait animation.
%
%
The resulting animations try to emulate natural head motions and eye movements, but struggle to transfer the rich phonetic structure of speech to lip movements.
This is due to limitations in capturing both long- and short-term speech-animation dependencies, as well as the fact that SFT alone falls short of explicitly guiding models to learn fine-grained phoneme-to-lip synchronization and context-aware head movements.
%
%
Therefore, existing approaches to speech-driven portrait animation struggle to offer a holistic solution to the aforementioned challenges. 
%
%
\par
With these considerations, we present \model~ --- a speech-to-portrait animation framework designed for realistic generation
. In particular, accurate lip synchronization with speech requires the model to focus on the low-level acoustic structure surrounding each spoken word.
Complementing this, natural head movements and expression changes are driven by prosodic stresses and the overall semantic meaning of speech.
To achieve both goals, we create a multi-level speech representation to condition the video generation process.
This representation captures speech features at multiple levels of granularity, ranging from global prosodic cues to fine-grained phonetic structure. 
We inject these multi-level features to different transformer blocks through \textit{level selectors}, which choose the appropriate contextual level for each block from among all available levels.
This enables our approach to generate accurate lip movements driven by the speech structure and yield realistic and plausible head motions driven by prosodic cues.
\par
Traditionally, RL-based generative modeling methods have relied on Proximal Policy Optimization (PPO) \cite{ren2024diffusion, wu2025reinforcement}, while more recent works have demonstrated strong performance using Group Relative Policy Optimization (GRPO) in vision tasks \cite{xue2025dancegrpo, wu2025reinforcement}. In parallel, Direct Preference Optimization (DPO) has emerged as an alternative paradigm for preference learning. However, these approaches typically rely on either human annotations~\cite{rafailov2023direct, liu2025videodpo} or pretrained reward models~\cite{ wu2023human,wang2025taming}.
Leveraging human annotations for preference learning with (DPO) is expensive and difficult to scale. On the other hand, designing custom reward functions requires strong, often ill-defined assumptions about what constitutes realism, and typically emphasizes only specific aspects of animation, potentially introducing bias toward those isolated factors.
Moreover, there is no single, holistic reward function that can reliably capture perceptual realism in generated videos; in practice, such definitions are inherently task-dependent and fail to generalize across diverse scenarios.

%
Our approach addresses these limitations by introducing a reward-free, RL-inspired fine-tuning strategy that improves perceptual realism and naturalness in generated videos. The model learns realism in a self-supervised manner, without relying on human annotations or explicitly defined reward functions. Instead, it constructs a curriculum of negative samples alongside ground-truth data, isolating key realism factors and enabling learning through direct comparisons between real and synthesized samples.
This formulation removes the need for both overly specified reward functions and costly annotation processes, relying instead on semantic differences between real data distributions and generated unrealistic samples.
\par
Combining the base generative framework with the proposed multi-level speech representation and reward-free RL training strategy yields natural talking portrait videos with improved speech-to-lip synchronization and high expressivity.


\section{Related Works}
\subsection{Diffusion-based Video Generation.}
Diffusion models have recently become the dominant paradigm for high-fidelity video generation. 
Early methods primarily leverage UNet architectures; for instance, AnimateDiff \cite{guo2023animatediff} inflates text-to-image models by incorporating temporal layers to capture motion dynamics, effectively exploiting pretrained image generators for spatial quality. Other approaches, such as Stable Video Diffusion \cite{blattmann2023stablevid}, are trained as dedicated video generation models. 
More recent approaches based on DiT architectures advance the field by integrating 3D VAEs with transformer-based sequence modeling \cite{gao2025wan, yang2024cogvideox}, enabling fine-grained control over identity, expressions, and motion. 
%
%
Parallel efforts in preference-aware generation, such as Diffusion-DPO \cite{wallace2024diffusion} and VideoDPO \cite{liu2025videodpo}, extend the DPO paradigm to align outputs with human judgments, optimize motion dynamics, and refine perceptual quality. 
While these methods demonstrate impressive video synthesis capabilities, they are largely agnostic to audio inputs and the specific challenges of speech-driven portrait animation. 
Our work builds on these advances by integrating multi-level speech conditioning and reward-free human preference alignment to generate realistic and expressive talking portraits.
\subsection{Audio-driven Portrait Animation.}
Audio-driven talking head generation focuses on synthesizing realistic facial motion from speech.
Early methods~\cite{zhou2020makelttalk, gururani2023space, ng2022learning} rely on 3D intermediate representations, such as facial landmarks and 3D Morphable Models, to guide lip synchronization, but often produce limited expressions and constrained head motions. 
Subsequent end-to-end approaches, including Wav2Lip \cite{wav2lip}, SadTalker \cite{sadtalker}, and FantasyTalking \cite{fantasy}, leverage audio-visual learning to improve lip alignment and overall expressivity.
Transformer- and diffusion-based frameworks, such as Hallo3 \cite{hallo3}, VASA-1 \cite{vasa}, and FantasyTalking\cite{fantasy}, further enhance realism by modeling temporal context and prosodic cues through cross-attention mechanisms to achieve precise audio-to-face alignment. 
Nevertheless, existing approaches face persistent challenges: maintaining fine-grained articulatory synchronization, capturing high-frequency facial dynamics that extend beyond the lip region, while modeling both local and global speech-motion correlations.
To this end, our method introduces multi-level speech representations and temporal motion modulation within a flow-matching framework, allowing concurrent lip-sync accuracy and natural expressive motion in high-fidelity portrait videos.

\section{Method}

 Our framework \model~ builds on the Wan 5B pretrained video generation model \cite{wan}. Similar to prior approaches such as fantasytalking (which uses Wan 14B) \cite{fantasy} and Hallo3  (based on CogVideoX-5B \cite{yang2024cogvideox}), we adopt models from a similar class.

%

\begin{figure}[t]
    \centering
  \includegraphics[width=0.6\textwidth]{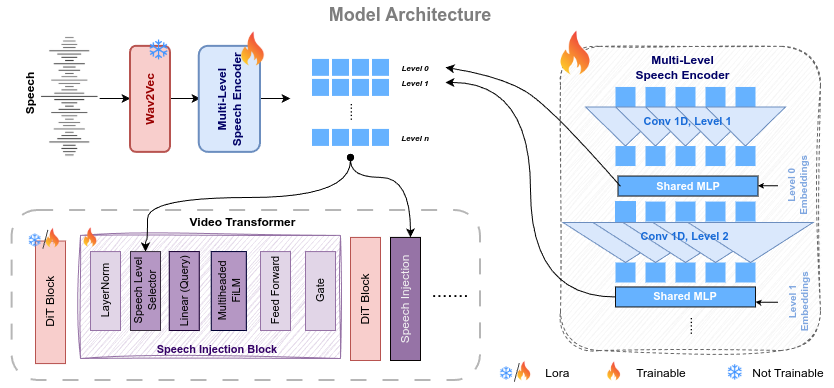}
  \caption{We present a speech-to-video generation framework that produces speech-synchronized, natural-looking portrait animations. The model incorporates multi-level speech guidance to jointly leverage local and global cues for richer neighborhood representations.
  }
  \label{fig:method1}
\end{figure}

%
 We introduce our multi-level speech representation in Sec. \ref{secMLG}, which is then injected into the DiT blocks through multi-head FiLM-based speech injection blocks (Sec. \ref{sec:audioinjection}).
At this stage, the resulting framework is trained with a flow-matching objective by applying low-rank adaptation (LoRA) on the pretrained DiT blocks and fully optimizing Speech Injection blocks and the multi-level speech encoder. We then introduce a second optimization stage based on the reward-free RL finetuning strategy (Sec. ~\ref{sec:noreward}), which enhances perceptual realism and natural speech-to-animation transfer. The complete framework is illustrated in Fig. \ref{fig:method1} and Fig \ref{fig:method2}.
%

\subsection{Multi-level speech Representation}
\label{secMLG}

We observe that the input speech signal inherently carries both global and local information. 
For example, when posing a question or saying 'no', humans tend to perform a preparatory gesture like head-shaking. Here, the head-turning motion begins before the speech is articulated. This implies that video frames preceding the onset of speech must anticipate and initiate the motion, meaning that future speech information is necessary for accurate generation (refer to Fig. \ref{fig:anticipatory-ablation}).
Relying solely on a simple frame-to-speech alignment can often constrain the naturalness and expressiveness of motion. 
%
%
Additionally, not all transformer layers require access to the full global context; some layers may focus on local features, while others capture global patterns or a combination of both. 
Motivated by these observations, we develop the proposed multi-level speech representation. 
\par
First, the speech input is processed by a multi-level encoder, as shown in Fig. \ref{fig:method1}, 
Each level consists of a CNN layer with kernel size $K_l$ and a shared MLP across all $L$ levels, which is additionally conditioned on an embedding for the level-index.
The kernel size $K_l$ increases from level 0 to $L$, effectively expanding the temporal context window from local to more global.
The resulting activations $\mathbf{f}_l$ from all the levels are concatenated and provided as input to each speech-injection block in the base video diffusion model.  
\subsection{Speech Injection}
\label{sec:audioinjection}
For each speech-injection block, the multi-level speech encoding is combined using a weighted gating mechanism:

\begin{equation}
\mathbf{A} = \sum_i^L \bigl[\text{softmax}(\mathbf{W})\bigr]_i \mathbf{L}_i
\end{equation}


where $\mathbf{L} \in \mathbb{R}^{L \times d}$ represents the outputs of the $L$ encoder levels, $\mathbf{W} \in \mathbb{R}^{L}$ are learnable selection weights, and $d$ is the feature dimension.
%
Next, we employ a multi-headed FiLM layer to condition the video generation model with the speech conditioning $\mathbf{A}$. 
%
%
%
Unlike approaches~\cite{fantasy} that use windowed cross-attention, this multi-head FiLM design provides more representational flexibility (refer to Tab. \ref{tab:abalation}, \ref{tab:hdtf}).
While each FiLM head learns shift and scale transformations, applying windowed cross-attention can be interpreted as only learning shifts, limiting its expressiveness.
Complete process in the multi-head film layer is presented in Algorithm~\ref{alg:mhfilm}.
Formally, each multi-head FiLM layer gets hidden states $\mathbf{H}_{in}$ of video tokens from the preceding layer, and speech conditioning $\mathbf{A}$.These hidden states of sequence length $S$ are projected into queries by a linear layer $\mathbf{W}_q$. and Per-head FiLM parameters $\gamma_k$ and $\beta_k$ are learned through a linear transformation $\mathbf{W}_c^{(k)}$ on speech conditioning.

\begin{algorithm}[t]
\caption{Multi-Head FiLM Layer}
\label{alg:mhfilm}
\small 

\begin{algorithmic}[1]
\STATE \textbf{Input:} $H_{\text{in}} \in \mathbb{R}^{S \times d_{\text{in}}}$, $\mathbf{A} \in \mathbb{R}^{L \times d}$, $K$
\STATE \textbf{Output:} $F_{\text{mod}} \in \mathbb{R}^{S \times d}$

\STATE $H_{\text{proj}} \leftarrow W_Q(H_{\text{in}})$ \hfill \# $(S \times d)$
\STATE $H, \mathbf{A'} \leftarrow \text{reshape\_to\_heads}(H_{\text{proj}}, \mathbf{A})$ 

\FOR{$k = 1$ \TO $K$}
    \STATE $(\gamma_k, \beta_k) \leftarrow \mathbf{W}_c^{(k)}(\mathbf{A'}_k); \mathbf{F}_k \leftarrow \gamma_k \odot H_k + \beta_k$ 
\ENDFOR

\STATE $\mathbf{F}_{\text{mod}} \leftarrow \text{Concat}(\mathbf{F}_1, \dots, \mathbf{F}_K)$
\end{algorithmic}
\end{algorithm}

\begin{figure}[t]
    \centering
  \includegraphics[width=1\textwidth]{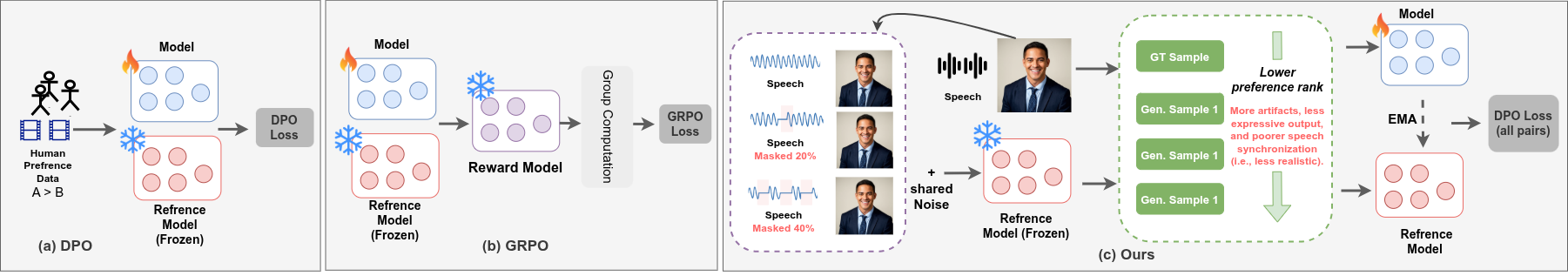}
  \caption{We present a novel reward-free reinforcement learning method based on a guided ranking strategy, which generates informative negative samples alongside real ones during training. Unlike GRPO (b), our approach does not require a reward model, and unlike DPO (a), it does not rely on human intervention to generate preference data.
  }
  \label{fig:method2}
\end{figure}
\subsection{No-Reward RL Fine-tuning}
\label{sec:noreward}
%
\subsubsection {Motivation}

While training our base framework with multi-level speech representation can improve animation quality,
%
The objective function still relies on the flow-matching objective to learn speech-to-animation transfer.
This objective provides no explicit guidance for distinguishing natural from unrealistic motion, as the trajectory-matching L2 loss operates pixel-wise and is agnostic to physical plausibility.
%
%
%
%
%
To address this, recent generative frameworks adopt reinforcement learning-driven finetuning methods such as direct preference optimization (DPO)~\cite{wallace2024diffusion, liu2025videodpo} or group relative policy optimization (GRPO)~\cite{xue2025dancegrpo, park2025deepvideo} for better physical plausibility or human preference alignment. These approaches utilize reward models or human annotations~\cite{cui2025hallo4} (as visualized in Fig.~\ref {fig:method2}) to perform preference optimization on the pretrained models.
%


%
However, large-scale human annotation becomes infeasible, and reward models rely on specific assumptions that introduce biases. 
%
%
More fundamentally, the core issues lie in the use of reward models: what kind of reward should be used? How can we formally define what constitutes a realistic generation or a physically plausible one through a scalar reward signal? 
%
These open questions turn this process into an ill-defined one.
For example, a naïve implementation might use SyncNet~\cite{syncnet} as a reward signal to improve speech-lip synchronization. However, this introduces a strong bias toward lip matching while neglecting other perceptual qualities such as expressiveness, image quality, head motion, and eyebrow movement. Comprehensively judging every such aspect is non-trivial, and even aggregating multiple pretrained reward models fails to capture the full picture, a task that would ultimately require a true world model, which does not yet exist.%
Considering these limitations, we instead introduce a novel approach that eliminates the need for both a reward model and human annotators. 
During the second-stage optimization, we teach the model realism by synthesizing auto-ranked negative samples that isolate the specific aspects the model needs to improve.

\subsubsection{The Emergence of Realism}
For each ground-truth (GT) sample, we synthesize $N$ progressively speech-masked samples to form the ordered set $\mathcal{S} = \{s_0, s_1, \dots, s_N\}$, where $s_0$ represents the original GT and $s_i$ denotes a synthesized sample with a higher masking ratio than $s_{i-1}$.  We adopt a random hard-masking strategy, using small windows scattered across the speech sequence. This produces outputs that are progressively less expressive and less synchronized, with increasing visual artifacts arising from the distribution gap between masked and unmasked speech. Crucially, the degree of degradation scales with the masking ratio: the more aggressively the speech is masked, the more unrealistic, unsynchronized, and artifact-prone the output becomes. By keeping the initial noise consistent across all synthetic samples, we can automatically rank them based on the masking percentage. These masked samples also serve as negative examples relative to the GT sample $s_0$. This approach allows us to automatically synthesize $N$ unrealistic samples and leverage the gap between the ground-truth (GT) sample and the synthetic samples to encourage the model to learn realistic features like image quality. Realism and physical plausibility emerge naturally from this discrepancy between the synthetic outputs and the GT data. This design isolates the effect of non-realism, giving the model a clear and focused signal to learn its significance.

We use the auto-ranked tuple of samples as guidance. Because they share the same initial noise, they are effectively identical samples with varying levels of realism degradation. We then apply DPO over all pairs $(s_i, s_j)$ with $i<j$, giving the total loss:
\begin{equation}
\begin{aligned}
\mathcal{L}_{\mathrm{DPO}}^{\mathrm{all}} = - \mathbb{E}_{t \sim \mathcal{U}(0,1)} \sum_{i<j} \log \sigma \Big( - \beta \omega_t [\Delta_\theta - \Delta_{\mathrm{ref}}] \Big), \\
\text{where } \Delta_u = \| v_i - u(s_i^t, t) \|^2 - \| v_j - u(s_j^t, t) \|^2.
\end{aligned}
\end{equation}
Here, $v_i, v_j$ are velocity targets for win/loss samples; $v_\theta, v_{\mathrm{ref}}$ are model and reference predictions; $\beta$ is temperature; and $\omega_t$ is a timestep weight.
We also update the reference model using an exponentially moving average (EMA), gradually shifting the distributions of both models toward more realistic generations. This enables the model to learn realism in a self-supervised manner, relying only on real speech–video pairs, without human annotations or reward models (Figure~\ref{fig:method2}).

%
%


\section{Experiments}

\textbf{Datasets}. 
Our training data consists of three datasets: Hallo3 \cite{hallo3}, CelebV-HQ \cite{celebv}, and the Seamless Interaction dataset \cite{seamless_interaction}.
%
%
Since the Seamless Interaction dataset consists of studio-quality footage captured in a controlled environment, its high audio clarity makes it valuable for learning accurate speech–lip correspondence. However, because it does not reflect in-the-wild conditions, we include it in the training set with a lower sampling probability.
For evaluation following prior methods, we incorporate around 200 clips from the HDTF dataset \cite{hdtf}, and for the in-the-wild setting, we test our model on a collection of in-the-wild samples from the Hallo3 dataset \cite{hallo3} comprising 50 randomly selected videos.

\noindent \textbf{Baselines.} We evaluate our model against ground-truth and recent state-of-the-art methods, including Hallo3~\cite{hallo3}, which is based on the CogVideoX~\cite{yang2024cogvideox} video generation backbone with cross-attention for speech conditioning; FantasyTalking~\cite{fantasy}, which leverages a larger Wan 14B~\cite{wan} model and additional facial encoding for speech-to-portrait animation; and Sonic ~\cite{sonic}, built on Stable Video Diffusion~\cite{blattmann2023stable} with audio injected via cross-attention mechanisms.


\begin{figure}[t]  
  \centering
  \includegraphics[width=0.95\linewidth]{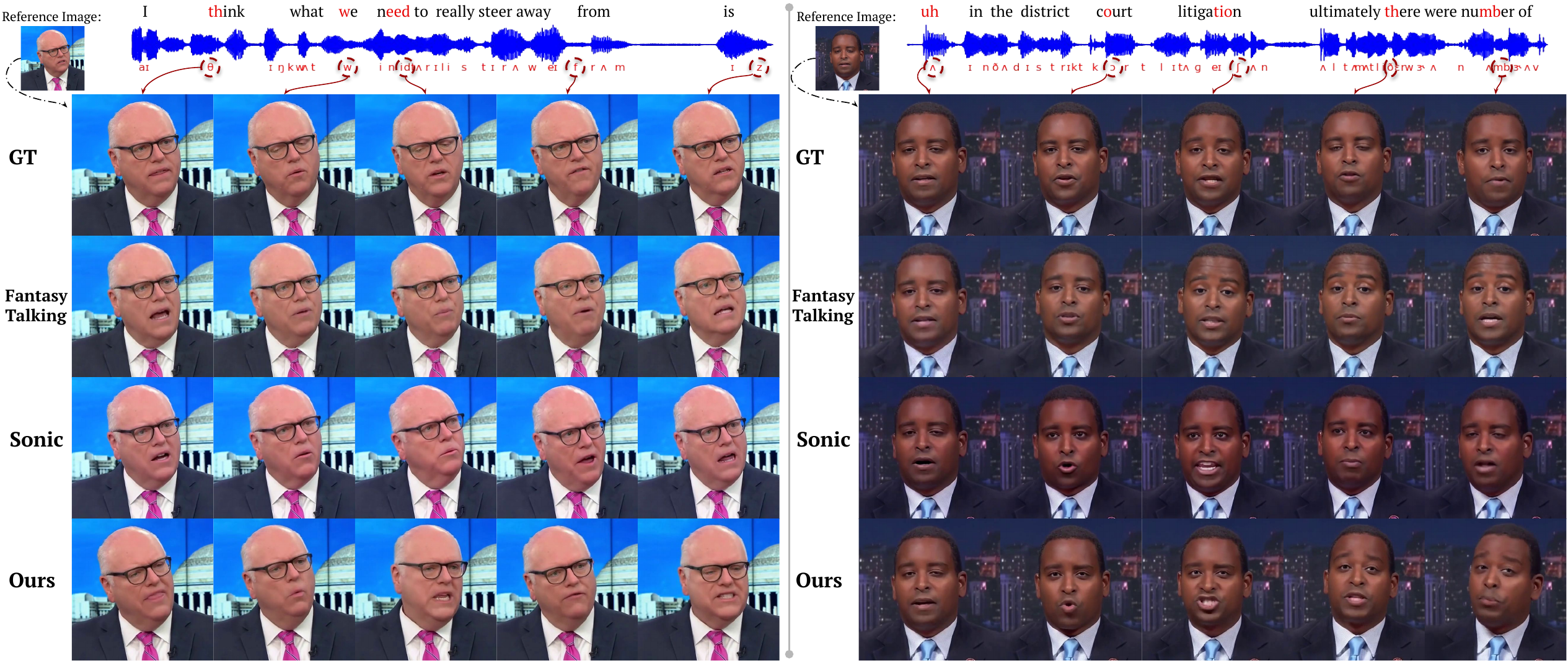}
  \caption{Qualitative results on HDTF dataset~\cite{hdtf}. Note the accurate phoneme-to-lip synchronization on phonemes like /$\theta$/ (think) and /f/ (from).}
  \label{fig:qualitative}
\end{figure}
\subsection{Perceptual Evaluation.}
\noindent \textbf{Qualitative Results.} 
We visualize samples from the HDTF dataset in Figure~\ref{fig:qualitative}, comparing our generations against competing methods and the Ground Truth (GT). 
Visually, we observe accurate phoneme-to-lip-shape matching, which demonstrates our model's superior lip-sync capability.
Furthermore, as demonstrated in Figure~\ref{fig:qualitative}, our model generates highly diverse and expressive head movements. This is particularly evident in Figure~\ref{fig:qualitative}, where our method exhibits noticeably richer and more varied facial expressions, particularly in head pose and eyebrow movements, compared to the more limited range of other methods.

%

\begin{figure}[t]  
  \centering
  \includegraphics[width=0.6\linewidth]{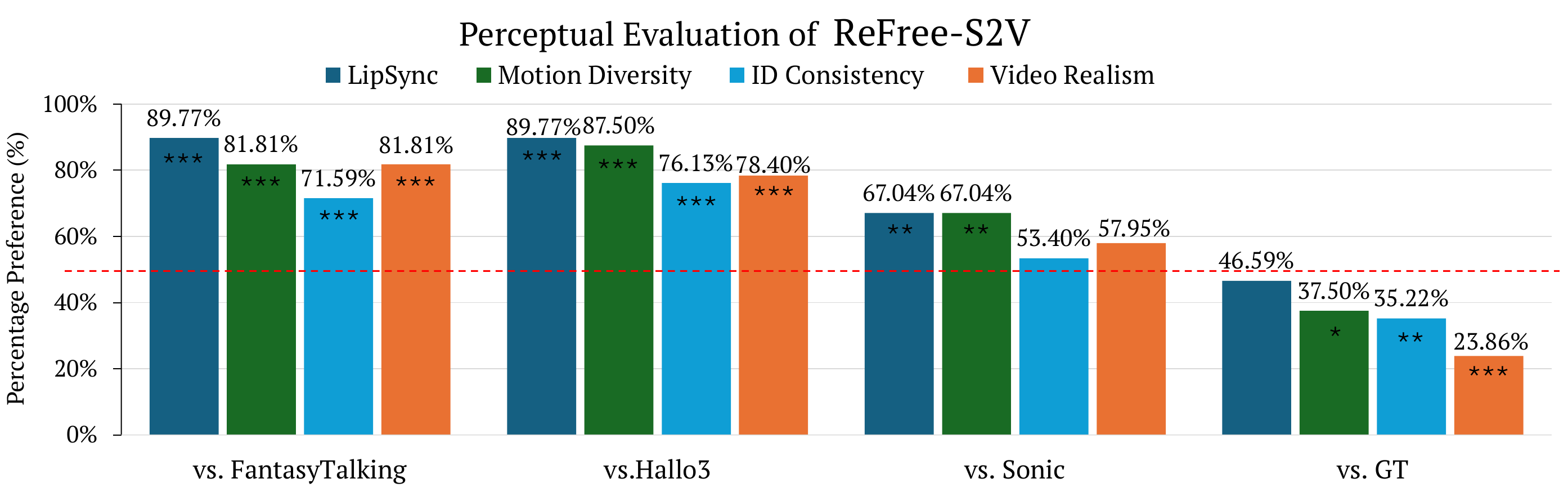}
  \caption{\textbf{Perceptual evaluation of ReFree-S2V.}
    We report pairwise preference percentages of our method against four baselines. The red line indicates the chance level (50\%). Here, *: (p < 0.05), **: (p < 0.01), and ***: (p < 0.001).}
  \label{fig:userstudy}
\end{figure}

\noindent \textbf{User Study.}
ReFree-S2V outperforms all baselines across all metrics, showing the most significant gains in Lip Synchronization and Video Realism. Notably, when compared to the ground truth (GT), our method achieves near-parity in lip synchrony while maintaining strong results in diversity and consistency. Furthermore, in terms of visual realism, our generated samples are preferred over the ground truth in approximately 24\% of cases.
%

\subsection{Quantitative Evaluation}
\textbf{Evaluation Metrics.} 
To assess motion–video alignment, we adopt the widely used Sync-C and Sync-D metrics~\cite{sync}, which measure the confidence and temporal distance of audio–visual synchronization. 
These metrics reflect how well generated human motions align with speech, particularly in highly dynamic segments.
For facial expression quality, we employ \emph{mesh-FID}. Specifically, we extract 3D facial meshes using MediaPipe\cite{mediapipe} and compute the Fréchet Inception Distance (FID)~\cite{fid} directly on the vertex representations without normalizing them into image space. This metric quantifies discrepancies between generated and ground-truth videos in terms of facial expressions, as well as head movement and pose.

To evaluate overall visual realism and alignment with the reference videos, we report both FID \cite{fid} and FVD~\cite{fvd}. 
Finally, to measure the degree of motion dynamics in the generated videos, we adopt the \emph{Dynamic Degree} metric~\cite{vbench}. This metric penalizes static videos with minimal motion, encouraging expressive results that include not only accurate lip movements but also coherent facial responses.

\begin{table*}[t]
  \centering

  \vspace{\baselineskip} 
  \caption{Performance comparison with existing approaches on the HDTF dataset.}
  \label{tab:hdtf}
  \vspace{\baselineskip} 
  
  \small
  \setlength{\tabcolsep}{10pt}
  \renewcommand{\arraystretch}{1.3}

  \begin{tabular}{@{} l S[detect-weight, table-format=1.3] S[detect-weight, table-format=2.3] S[detect-weight, table-format=2.3] S[detect-weight, table-format=3.3] S[detect-weight, table-format=1.4] S[detect-weight, table-format=1.3] @{}}
    \toprule
    \textbf{Method} & {Sync-C $\uparrow$} & {Sync-D $\downarrow$} & {FID $\downarrow$} & {FVD $\downarrow$} & {Mesh-FID $\downarrow$} & {Dyn. Deg. $\uparrow$} \\
    \midrule
    Sonic          & 7.406          & 7.444           & 13.242          & 72.142          & 0.0507          & 0.238 \\
    FantasyTalking & 3.518          & 11.072          & 16.488          & 122.167         & 0.1094          & 0.021 \\
    Hallo3         & 5.949          & 9.181           & 14.656          & \bfseries 70.372 & 0.0336          & 0.516 \\
    \midrule 
    Ours$_{\text{w/o RL}}$  & 7.687          & 7.160           & 12.319          & 88.015          & 0.0290          & 0.534 \\
    \textbf{Ours}   & \bfseries 7.999 & \bfseries 7.011 & \bfseries 11.643 & 86.182          & \bfseries 0.0099 & \bfseries 0.559 \\
    \bottomrule
  \end{tabular}

  \vspace{\baselineskip} 
\end{table*}

\begin{table*}[t]
  \centering
  
  \vspace{\baselineskip} 
  \caption{Performance comparison with existing approaches on the in-the-wild samples from the Hallo3 dataset.}
  \label{tab:wild}
  \vspace{\baselineskip} 

  \small
  \setlength{\tabcolsep}{10pt}
  \renewcommand{\arraystretch}{1.3}

  \begin{tabular}{@{} l S[table-format=1.3] S[table-format=2.3] S[table-format=2.3] S[table-format=3.3] S[table-format=1.4] S[table-format=1.2] @{}}
    \toprule
    \textbf{Method} & {Sync-C $\uparrow$} & {Sync-D $\downarrow$} & {FID $\downarrow$} & {FVD $\downarrow$} & {Mesh-FID $\downarrow$} & {Dyn. Deg. $\uparrow$} \\
    \midrule
    Sonic          & 5.990          & 8.224           & 34.979          & 270.047          & 0.1340          & 0.22 \\
    FantasyTalking & 3.001          & 11.294          & 37.877          & 277.888          & 0.2249          & 0.02 \\
    Hallo3         & 5.041          & 9.669           & 38.762          & 281.060          & 0.1204          & 0.36 \\
    \midrule 
    Ours$_{\text{w/o RL}}$  & 6.064          & 8.096           & 36.921          & 277.768          & {\bfseries 0.1017} & {\bfseries 0.76} \\
    \textbf{Ours}   & {\bfseries 6.651} & {\bfseries 7.781} & {\bfseries 33.345} & {\bfseries 258.755} & 0.1304          & 0.58 \\
    \bottomrule
  \end{tabular}

  \vspace{\baselineskip} 
\end{table*}

\noindent \textbf{Setup}. To ensure fair comparison across different methods, we resize all videos to the lowest common resolution, 480×480, and standardize the frame rate to 15 FPS. We evaluate on the standard HDTF dataset \cite{hdtf}. Since HDTF is generally considered an easier benchmark, we also follow prior work by evaluating on a challenging “in-the-wild” subset from the hallo3 dataset.

As shown in Table~\ref{tab:hdtf}, our method establishes new state-of-the-art performance across most metrics. Notably, we achieve the lowest FID, the best Sync-C and Sync-D scores, and the largest improvement in dynamic degree, demonstrating superior lip synchronization, diversity, expressiveness, and overall fidelity to ground truth. The only exception is FVD, where we remain highly competitive. In the unconstrained "in-the-wild" setting (Table~\ref{tab:wild}), our method also establishes new state-of-the-art performance across all metrics. Notably, our RL-based variant achieves the best scores for FID, FVD, Sync-C, and Sync-D, while the non-RL variant demonstrates the highest dynamic degree and Mesh-FID. We attribute the RL version's slightly lower dynamic degree (compared to the non-RL version) to the training process: RL effectively dampens unrealistic, explosive motions, grounding the output in realism. This hypothesis is supported by the RL version's superior FID and FVD scores. To summarize, our method sets a new benchmark across both settings, delivering significant improvements in realistic, expressive, and synchronized video generation.

\subsection{Ablative Analysis.}

\subsubsection{RL-based Training.}
  


\begin{table}[H]
  \centering

  \vspace{\baselineskip} 
  \caption{Component-wise ablation: synchronization metrics across architectural variations. Bold indicates best.}
  \label{tab:abalation}
  \vspace{\baselineskip} 
  
  \small
  \setlength{\tabcolsep}{10pt}    
  \renewcommand{\arraystretch}{1.3} 

  \begin{tabular}{@{} l S[detect-weight, table-format=1.3] S[detect-weight, table-format=2.3] @{}}
    \toprule
    \textbf{Method Configuration} & {Sync-C $\uparrow$} & {Sync-D $\downarrow$} \\
    \midrule
    \textbf{Ours (Full)}                & \bfseries 1.721 & 12.505 \\
    w/o Multi-headed FiLM (Cross-Attn)  & 1.653           & \bfseries 12.476 \\
    w/o Multi-Level Embedding           & 1.405           & 12.740 \\
    \bottomrule
  \end{tabular}

  \vspace{\baselineskip} 
\end{table}
We fine-tune our supervised model on a dataset of approximately 20,000 samples randomly selected from the \cite{celebv} dataset, each consisting of a tuple of (ground truth, generated sample 0 (0\% masked), generated sample 1 (20\% masked), generated sample 2 (40\% masked)) data points, as described in Sec.~\ref{sec:noreward}. The model is trained for roughly 5,000 steps with an effective batch size of 8, using a beta value of 2,500 and a learning rate of $1 \times 10^{-5}$ with AdamW as the optimizer. Additionally, we apply an exponential moving average (EMA) to the reference model, which allows the model to gradually shift toward a better distribution.

As shown in Tables~\ref{tab:hdtf} and~\ref{tab:wild}, our RL-enhanced model achieves state-of-the-art performance across most metrics. Compared to the non-RL variant, we observe that reinforcement learning markedly improves speech synchrony, as indicated by higher Sync-C and Sync-D scores, as well as better FID and FVD values. We attribute these gains to the RL training, which effectively enhances both the realism and the audio-visual alignment in speech-to-video generation. We also provide an ablation on different values for the beta in Table \ref{tab:beta_eval}

\subsubsection{Component-wise Ablation}
\label{abl:anticipatory-motion}
\begin{figure}[t]  
  \centering
  \includegraphics[width=0.4\linewidth]{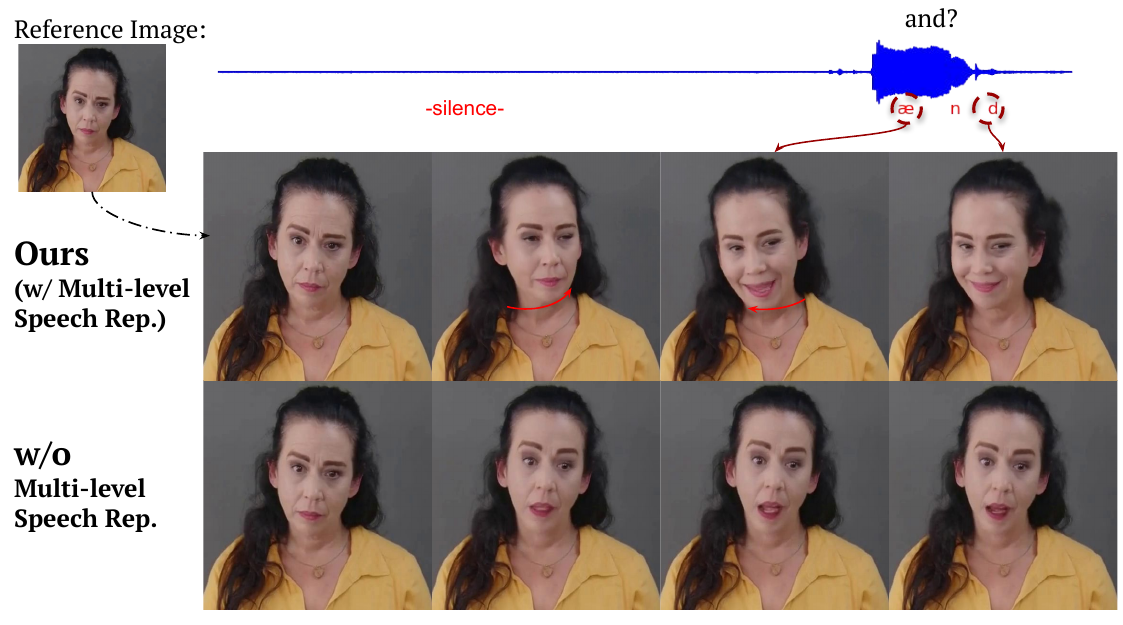}
\caption{Effect of using multi-level speech representations on capturing long-range future dependencies. With multi-level guidance, the model successfully initiates the motion sequence well before the corresponding speech stimulus, whereas the model without multi-level guidance fails to anticipate this future motion.}
  \label{fig:anticipatory-ablation}
\end{figure}

We evaluate the impact of capturing multiple views of the audio signal by employing a multilevel speech encoder that encodes different temporal neighborhood sizes at each level. For each speech-injection block, we subsequently use a level-selection layer to choose the most appropriate neighborhood scale. To assess the effectiveness of this design, we compare it against a baseline that uses a speech encoder without multi-level embedding. As shown in Table \ref{tab:abalation}, removing the multilevel embedding and selection mechanism leads to a noticeable drop in performance. We also visualize this effect in Figure \ref{fig:anticipatory-ablation}, where the multilevel embedding helps initiate an anticipatory head motion in response to speech stimuli far into the future, while the version without fails to capture such a relation.

We evaluate the effect of using a Multi-headed FiLM layer as opposed to cross-attention for speech-to-video frame conditioning (Table \ref{tab:abalation}). In both settings, speech tokens attend to video frames; however, replacing Multi-headed FiLM with cross-attention results in a drop in performance. This result suggests that in a windowed conditioning setting, Multi-headed FiLM is more effective, as it provides both scale and shift modulation of the visual features. In contrast, cross-attention can be interpreted as primarily inducing a shift, lacking the explicit multiplicative scaling that Multi-headed FiLM offers.




\begin{table*}[t]
  \centering
  \caption{Ablation on the value of beta in No-Reward RL training.}
  \label{tab:beta_eval}
  
  \small
  \setlength{\tabcolsep}{10pt}    
  \renewcommand{\arraystretch}{1.3} 

  \begin{tabular}{@{} l S[detect-weight, table-format=1.3] S[detect-weight, table-format=1.3] S[detect-weight, table-format=2.3] S[detect-weight, table-format=3.3] S[detect-weight, table-format=1.4] S[detect-weight, table-format=1.2] @{}}
    \toprule
    \textbf{Beta Value} & {Sync-C $\uparrow$} & {Sync-D $\downarrow$} & {FID $\downarrow$} & {FVD $\downarrow$} & {Mesh-FID $\downarrow$} & {Dyn. Deg. $\uparrow$} \\
    \midrule
    2.5k  & \bfseries 6.671 & 7.952          & 42.207          & 362.084          & \bfseries 0.1922 & \bfseries 0.84 \\
    5k & 6.665          & \bfseries 7.904 & 43.962          & 307.885          & 0.2608          & 0.68          \\
    7.5k & 6.527          & 8.155          & \bfseries 40.780 & \bfseries 270.773 & 0.3232          & 0.60          \\
    \bottomrule
  \end{tabular}
\end{table*}

\section{Conclusion} 
We introduced ReFree-S2V, a speech-to-video framework that captures expressive speech representations through multi-level guidance. Our approach improves speech–lip synchrony and visual realism via a novel reward-free reinforcement learning scheme, eliminating the need for human annotators and avoiding the biases and under-representation of realism often present in learned reward models. 
Both quantitative and qualitative experiments show that ReFree-S2V sets a new state of the art across standard metrics, outperforming previous methods. Our user study further demonstrates that ReFree-S2V achieves near–ground-truth speech synchrony and delivers significant improvements in perceived realism.
\bibliographystyle{unsrtnat} 
\bibliography{bibliography}

\small


\appendix

\appendix  
\clearpage
\section{Technical appendices and supplementary material}

\subsection{Computational Efficiency}

We compare the computational efficiency of our method against recent state-of-the-art approaches for audio-driven portrait animation in Table \ref{tab:efficiency}.
\begin{table}[H]
  \centering
  \caption{Computational efficiency comparison.}
  \label{tab:efficiency}

  \small
  \setlength{\tabcolsep}{10pt}
  \renewcommand{\arraystretch}{1.3}

  \begin{tabular}{@{} l S[detect-weight, table-format=2.1] S[detect-weight, table-format=1.1] @{}}
    \toprule
    \textbf{Method} & {VRAM (GB) $\downarrow$} & {FPS $\uparrow$} \\
    \midrule
    Sonic          &  13.8 & 0.8            \\
    FantasyTalking & 36.1           & 0.2            \\
    Hallo3         & 48.3           &  0.1  \\
    Ours           & 30.6           &  0.7  \\
    \bottomrule
  \end{tabular}
\end{table}




\subsection{Additional Qualitative results}

\begin{figure}[H]  
  \centering
  \includegraphics[width=\linewidth]{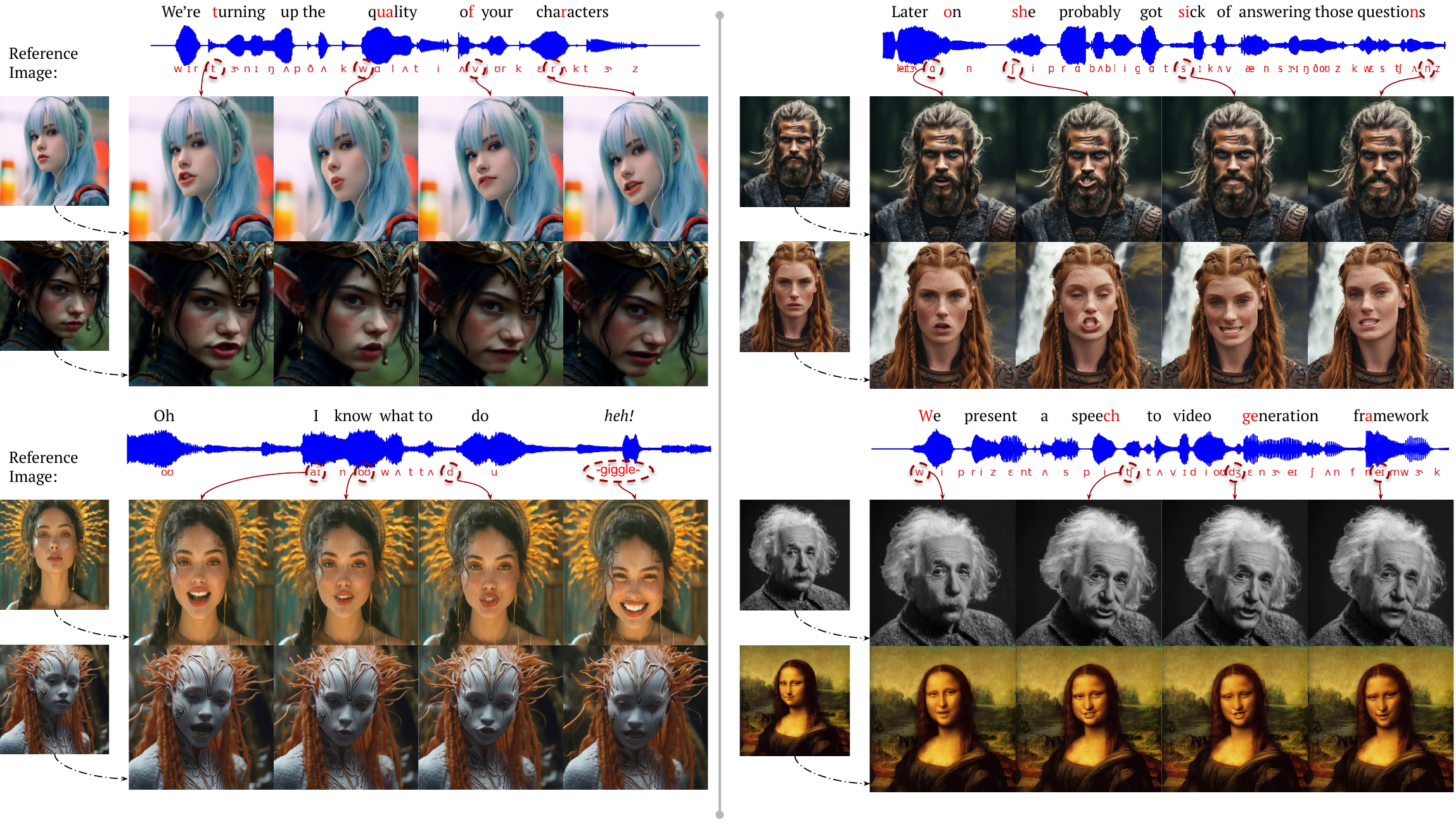}
  \caption{Qualitative results on in-the-wild AI-generated images. Note the accurate lip shapes for individual phonemes and expressive facial expressions for non-speech vocalizations like ``giggles''.}
  \label{fig:qualitativeall}
\end{figure}

\newpage
\subsection{Additional Quantitative  results}
We provide additional evaluation on the “in-the-wild” subset in Table~\ref{tab:fantasytalking_r4}. Note that FantasyTalking does not provide code for its evaluation metrics, so the metrics were re-implemented in our work.
We observe the best performance in Subject Dynamics and Body Dynamics, indicating improved expressiveness. We also observe a slight drop in ID-Consistency, which we attribute to the use of DINO features in the metric computation. Since DINO features are not identity-aware, more dynamic and expressive motions can lead to larger feature variations across frames, resulting in lower consistency scores. while we see almost no difference in the aesthetic score across methods.

\begin{table}[H]
  \centering
  \caption{Evaluation on the "in-the-wild" subset comparing FantasyTalking metrics.}
  \label{tab:fantasytalking_r4}

  \small
  \setlength{\tabcolsep}{10pt}
  \renewcommand{\arraystretch}{1.3}

  \begin{tabular}{@{} l S[detect-weight, table-format=1.3] S[detect-weight, table-format=1.3] S[detect-weight, table-format=1.3] S[detect-weight, table-format=1.3] @{}}
    \toprule
    \textbf{Method} & {ID-Consistency $\uparrow$} & {Subj.\ Dynamics $\uparrow$} & {BG Dynamics $\uparrow$} & {Aesthetic $\uparrow$} \\
    \midrule
    Sonic          & \bfseries 0.952 & 1.812           & 0.950           & 5.098          \\
    FantasyTalking & 0.941           & 0.896           & 0.745           & \bfseries 5.169 \\
    Hallo3         & 0.938           & 1.755           & 0.760           & 5.058          \\
    Ours           & 0.926           & \bfseries 3.426 & \bfseries 2.698 & 5.133          \\
    \bottomrule
  \end{tabular}
\end{table}

\subsection{Social Risks and Potential Mitigation}

Video generation frameworks carry social risks, such as misuse for deceptive content and privacy violations. These concerns underscore the importance of ethical guidelines, informed consent, and transparent disclosure of AI-generated content. Given the inclusion of personal identities, careful handling and robust watermarking are essential to distinguish AI-generated from real content.

\subsection{Limitations}
Although our method achieves improved realism, it is inherently limited by the underlying video generation model, resulting in high runtime and VRAM usage. 

A promising direction for future work is to improve efficiency via faster inference strategies. Techniques such as few-step distillation and efficient sampling could significantly reduce the number of required steps, enabling faster generation while preserving visual quality.

\subsection{Additional Implementation Details}
Our full training pipeline was executed on a cluster of eight NVIDIA H100 GPUs. For the Supervised Fine-Tuning (SFT) phase, we trained the model for approximately 50,000 steps using the AdamW optimizer. The learning rate was set to $1 \times 10^{-4}$ for the initial 30,000 steps and subsequently decayed to $1 \times 10^{-5}$ for the remaining 20,000 steps. 

To improve the fidelity of facial dynamics, we employed regional loss weighting to provide a stronger $L_2$ loss guidance signal. Specifically, we applied a weight of 100 to the lip region and 30 to the rest of the face, while the remainder of the video frames maintained a baseline weight of 1.

\subsection{User study}
\begin{figure}[H]  
  \centering
  \includegraphics[width=\linewidth]{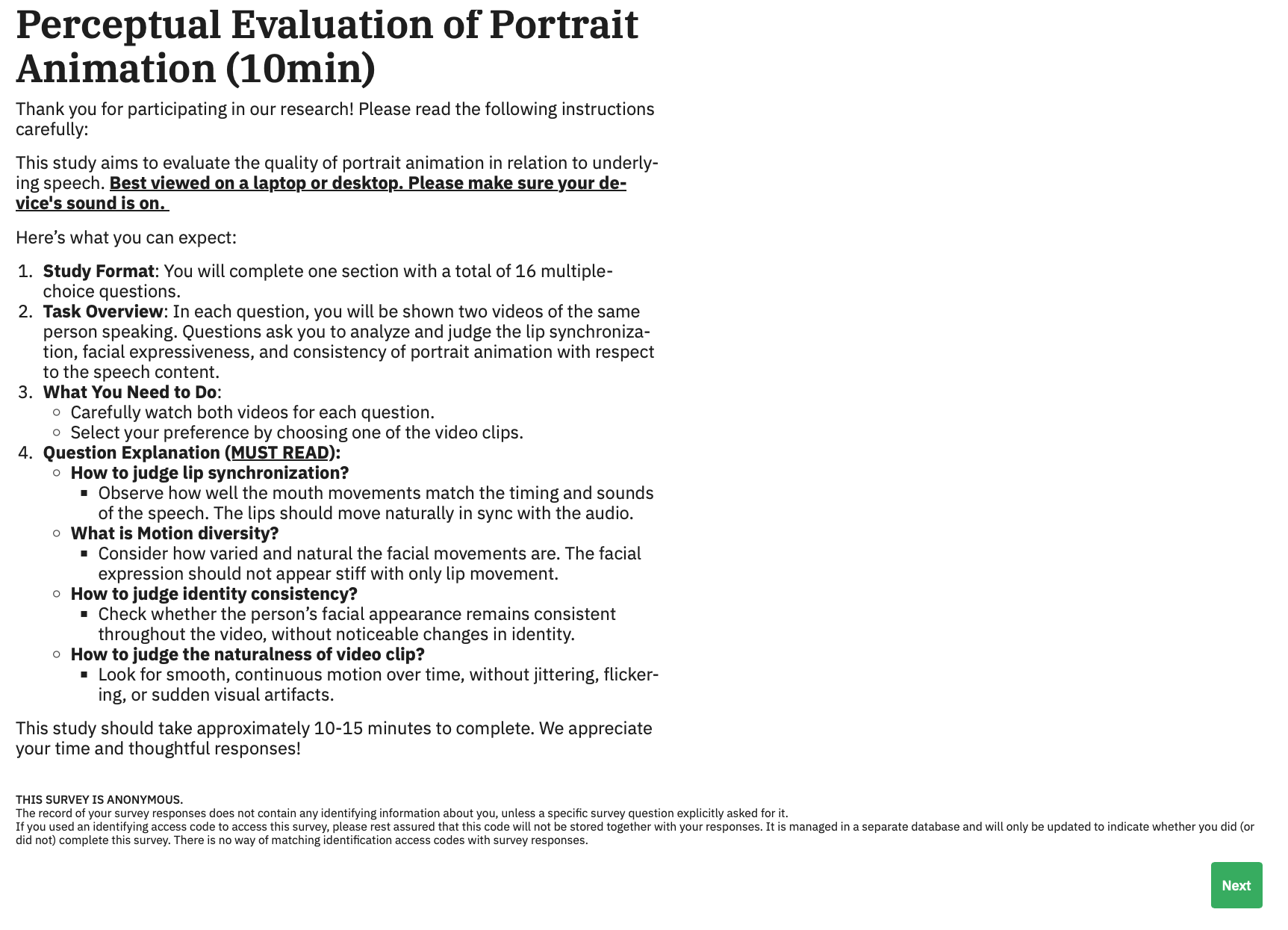}
  \caption{Screenshot of the user study}
  \label{fig:qualitativeall}
\end{figure}

\begin{figure}[H]  
  \centering
  \includegraphics[width=\linewidth]{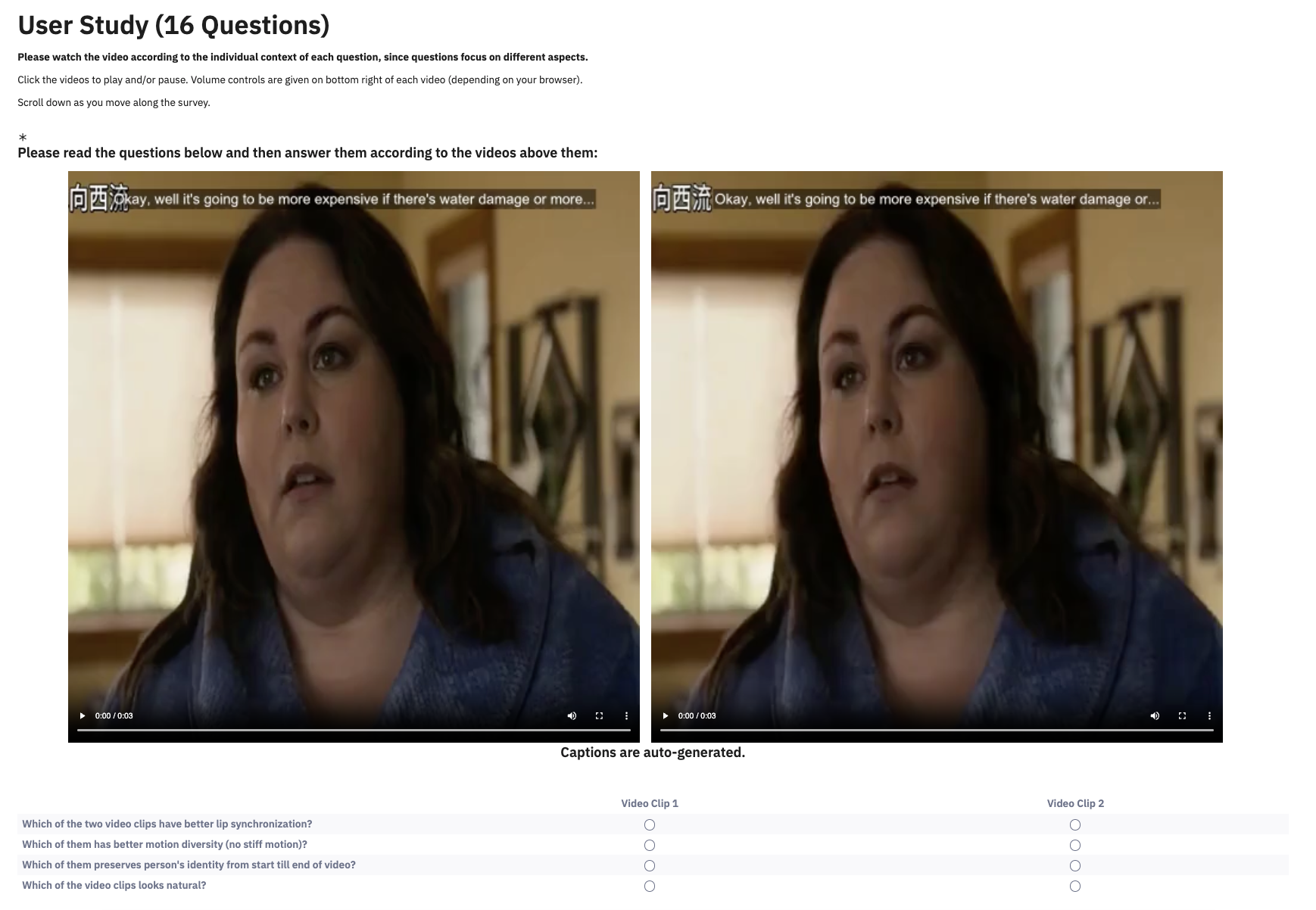}
  \caption{Screenshot of the user study}
  \label{fig:qualitativeall}
\end{figure}

\begin{figure}[H]  
  \centering
  \includegraphics[width=\linewidth]{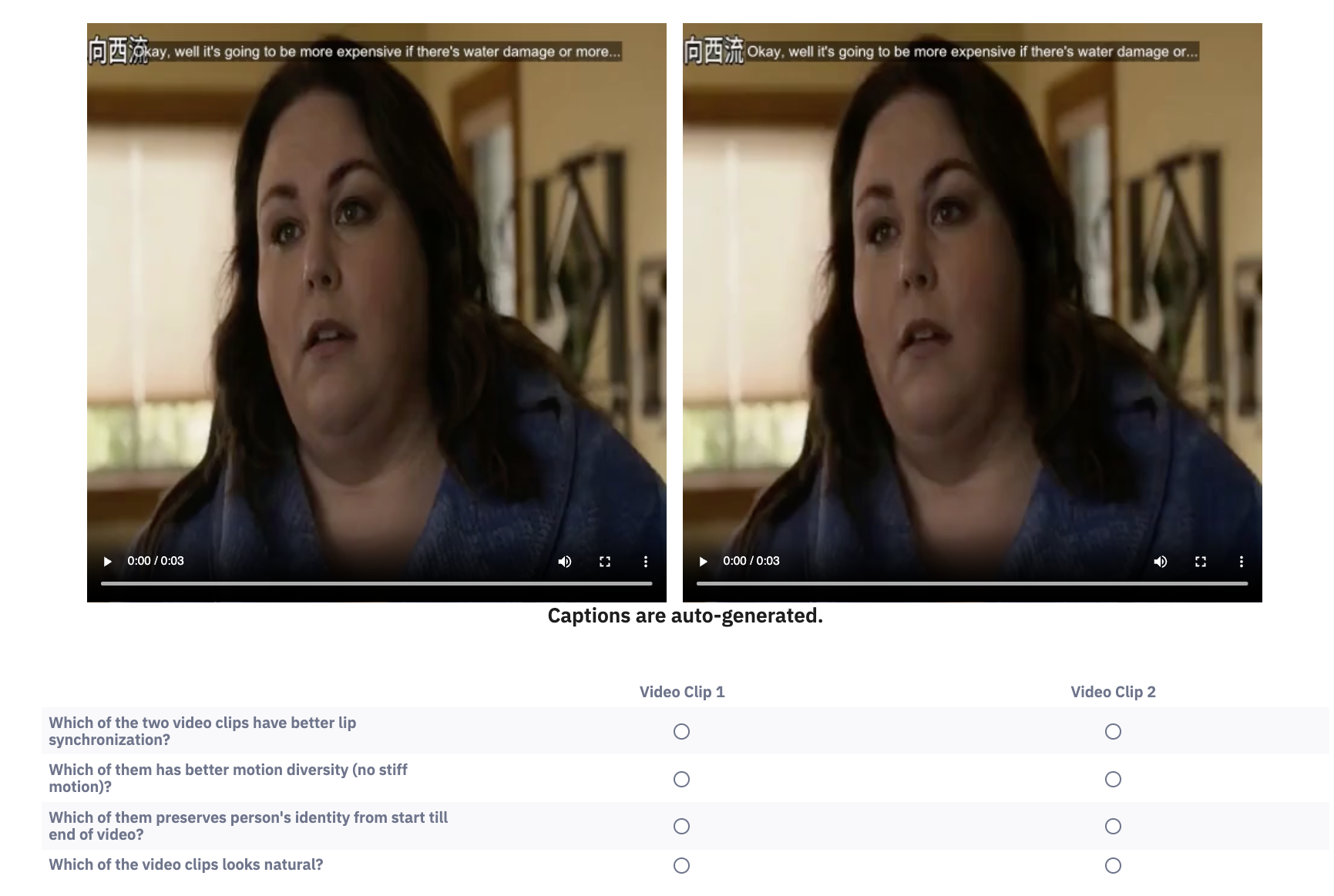}
  \caption{Screenshot of the user study}
  \label{fig:qualitativeall}
\end{figure}

\subsection{Additional License Information}

In this work, we use several open-source datasets and pretrained models. While all resources are properly cited throughout the paper, we provide here a non-exhaustive summary of the corresponding licenses and usage terms.

The Hallo3 dataset~\cite{hallo3} is released under the CC BY-NC-ND 4.0 license.  
HDTF~\cite{hdtf} is distributed under the GNU GPL v3.0 license.  
The Wan model~\cite{wan} is released under the Apache 2.0 license.  

The CelebV-HQ dataset~\cite{celebv} is available for non-commercial research purposes only.  

The Seamless Interaction dataset~\cite{seamless_interaction} is released under the CC BY-NC 4.0 license.  

Finally, some initial frame images used in the paper and supplementary videos were obtained from free-to-use resources requiring author attribution. These images were sourced from \texttt{freepik.com} and \texttt{vecteezy.com}, including work by Adam Zubek-Nizol.




\newpage

\end{document}